# A Learning-based Framework for Two-Dimensional Vehicle Maneuver Prediction Over V2V Networks

Hossein Nourkhiz Mahjoub, Amin Tahmasbi-Sarvestani, Hadi Kazemi, Yaser P. Fallah

*Abstract*— Situational awareness in vehicular networks could be substantially improved utilizing reliable trajectory prediction methods. More precise situational awareness, in turn, results in notably better performance of critical safety applications, such as Forward Collision Warning (FCW), as well as comfort applications like Cooperative Adaptive Cruise Control (CACC). Therefore, vehicle trajectory prediction problem needs to be deeply investigated in order to come up with an end to end framework with enough precision required by the safety applications' controllers. This problem has been tackled in the literature using different methods. However, machine learning, which is a promising and emerging field with remarkable potential for time series prediction, has not been explored enough for this purpose. In this paper, a two-layer neural network-based system is developed which predicts the future values of vehicle parameters, such as velocity, acceleration, and yaw rate, in the first layer and then predicts the two-dimensional, i.e. longitudinal and lateral, trajectory points based on the first layer's outputs. The performance of the proposed framework has been evaluated in realistic cut-in scenarios from Safety Pilot Model Deployment (SPMD) dataset and the results show a noticeable improvement in the prediction accuracy in comparison with the kinematics model which is the dominant employed model by the automotive industry. Both ideal and non-ideal communication circumstances have been investigated for our system evaluation. For non-ideal case, an estimation step is included in the framework before the parameter prediction block to handle the drawbacks of packet drops or sensor failures and reconstruct the time series of vehicle parameters at a desirable frequency.

## I. Introduction

Different mechanisms are proposed to achieve a higher level of safety in transportation systems via collision prevention or collision aftereffects reduction. The proposed safety strategies could be categorized in two main groups, namely passive and active safety systems. The former aims to decrease the severity of possible injuries to the vehicle passengers or vulnerable road users (VRUs) after collision occurrence, while the latter focuses on collision avoidance as its primary mission. Specific bumpers designed to preserve pedestrians' safety, airbags and seatbelts are some of the well-known examples of the passive safety mechanisms. Safety applications, such as forward collision warning (FCW) [1]-[3] , lane keep assistance (LKA) [4]-[6], automatic braking [7], adaptive cruise control (ACC) [8], [9] and cooperative pedestrian safety [10]-[12], are some of the most important active safety systems. The mentioned critical active safety applications in addition to some other applications, such as glare reduction [13] and fuel consumption optimization [14], [15], which are categorized under the more general title of Advanced Driver Assistance Systems (ADAS), are promising tools for realization of Intelligent Transportation System (ITS) notion.

In general, based on the available sources of information, ADAS could be classified into two main subclasses. Non-cooperative ADAS technology only relies on the local sensors, such as radars, cameras, or LiDARs, as its information source. However, the next level of ADAS, i.e. cooperative ADAS, incorporates Vehicle-to-Vehicle (V2V) and Vehicle-to-Infrastructure (V2I) communications as the additional resources to achieve a more precise situational awareness. This V2X communication could be attained by the virtue of Dedicated Short Range Communication (DSRC) standard [16] as a promising technology.

Situational awareness is a mandatory part of almost all ITS applications which enables them to have a better insight into their surrounding environment and consequently helps them to make wiser decisions. Precise prediction and modeling of driver behaviors significantly increase the level of situational awareness and empower the applications to have a clearer perception of the future critical moments of driving scenario. Therefore, driver behavior modeling has received a great deal of attention in ADAS design literature [17]-[19].

Vehicle trajectory is one of the best representatives of the driver behavior which is directly and immediately affected by different driver decisions. Hence, trajectory prediction is one of the most common ways for driver behavior modeling in the literature [20], [21]. In this work, we propose a novel trajectory prediction framework based on the machine learning methods. More specifically, we have two parallel trajectory predictors for longitudinal and lateral directions which could be incorporated separately or jointly into different types of ITS applications.

This material is based on work supported in part by the National Science Foundation under CAREER Grant 1664968 and in part by the Qatar National Research Fund Project NPRP 8-1531-2-651.

Hossein Nourkhiz Mahjoub is a PhD student with the Department of Electrical and Computer Engineering, University of Central Florida, Orlando, FL 32826 USA (e-mail: hnmahjoub@knights.ucf.edu).

Amin Tahmasbi-Sarvestani and Hadi Kazemi are PhD students with the Department of Electrical Engineering and Computer Science, West Virginia University, Morgantown, WV 26506 USA (e-mails: amtahmasbi@mix.wvu.edu, hakazemi@mix.wvu.edu).

Yaser P. Fallah is an associate professor with the Department of Electrical and Computer Engineering, University of Central Florida, Orlando, FL 32826 USA (e-mail: yaser.fallah@ucf.edu).

To realize driver's intention which leads us to predicting the vehicle trajectory, it is necessary to find some reliant methods which be able to differentiate between partially similar time series of vehicle parameters which belong to different high-level driving maneuvers. Support Vector Machine (SVM), Neural Networks (NN), Hidden Markov Models (HMMs) and Dynamic Bayesian Networks (DBNs), are some of the most frequently adopted methods in the literature for this purpose [21]-[32] . The set of vehicle parameters includes, but is not limited to, steering wheel angle, throttle, brake pedal position, lateral position, longitudinal position, velocity, acceleration, heading, yaw rate, and signaling status, which are accessible both from local sensors and DSRC communicated basic safety messages (BSMs) defined by the SAE J2735 standard [33].

Sensory data from side warning assist radars, head tracking cameras, ACC, and lane departure warning camera are used to construct a features vector in [21]. Then, this features vector is fed into Relevance Vector Machine (RVM), an extended version of SVM, to discriminate between lane keeping and lane change maneuvers.

Lane change intention tried to be recognized using SVM in [22] in a realistic driving data. The small number of resulting false alarms supports the applicability of SVM for this problem.

Authors in [23] proposed an SVM-based classifier to differentiate drivers' lateral maneuvers, such as lane change, via detection of preliminary behaviors, vehicle dynamics, and the environmental data before and during the maneuver. Three separate classifiers are utilized in this work for three consecutive maneuver stages, namely environmental intent, lateral intent, and lateral action.

Observed scenes of the host vehicle were classified using a hierarchical classifier proposed in [24]. The main goal of this work was the prediction of remote vehicles' driver maneuvers at a highway entrance with a mandatory lane change within 3 seconds.

To find out the possibility of a remote vehicles' lane change detection by an autonomous vehicle, authors in [25] developed a feed forward artificial neural network (ANN) which tries to predict remote vehicle trajectory using its movement history. Their results show that feed-forward ANNs using locally sensory data is not strong enough to predict a sufficiently accurate short term or long term trajectory.

An Object-Oriented Bayesian Network (OOBN) is proposed in [26] to detect different driving maneuvers in highways. These maneuvers are modeled as vehicle-lane and vehicle-vehicle relations on four hierarchical levels in this work. Vehicle-lane and vehicle-lane-marking relations are used to model a potential lane change and the lane marking crossing likelihood, respectively. In addition, vehicle-vehicle relations utilized to specify all possible maneuvers for a pair of vehicles.

A finite set of driving behaviors are classified and future trajectories of the vehicle are predicted based on currently understood situational context using a filter that was designed utilizing a DBN-based model [27]. The situational context was translated to the awareness of vehicle interactions with other traffic participants.

A major group of works in the literature rely on HMM technique to find the best maneuver which represents the observed vehicle parameters time series. For instance, [28] and [29], which are among the pioneer HMM-based works, have decomposed driver behaviors into two main categories of small scale and large scale behaviors. They assumed a Markovian property for the sequence of large-scale actions and suggested HMM to unveil the most probable next action. This claim was assessed and validated for lane change maneuver prediction. In addition, a Kalman filter is employed in [28] to estimate small scale changes inside each large-scale state.

An asymmetric Coupled HMM (CHHM) is utilized in [30] to model seven driver maneuvers based on inter-vehicle signals and cameras empirical information gathered for this purpose. Based on the provided results, CHMM, which is capable of modeling the interactions of different processes, seems to be a promising tool for situations in which HMM is not a perfect and complete solution, e.g., driver-environment interactions modeling.

Two types of lane changes, i.e. risky and safe, are detected and classified in [31] using a HMM, which tries to jointly model the vehicle dynamics sequence and driver gaze. The subjective risk level for HMM training in this work is obtained from a subset of available scenarios. The utilized performance criteria for model assessment, are the subjective scores in addition to the highest correlation between cumulative HMM log-likelihood ratios, which is defined as the ratio of the safe state likelihood to the risky state likelihood.

In another work, an HMM-based system is trained to identify the driver intention before performing a maneuver [32]. Inter-vehicle and environment perception data are utilized for this purpose. Different maneuvers are discriminated based on their likelihoods by HMM as a classifier. To evaluate the proposed approach, steering angle and yaw rate information are used for model training and then the trained model is tested for lane change intention detection.

In this paper, we propose a combination of nonlinear NN-based autoregressive (NAR), NN-based nonlinear autoregressive exogenous (NARX), and recurrent neural networks (RNN) to predict longitudinal and lateral vehicle trajectories, separately. The superiority of this method over the widely adopted kinematics model is validated using a set of realistic cut-in scenarios. Moreover, the effect of non-ideal communication, which is modeled in terms of packet drop rate, on our prediction framework performance is assessed.

The overall proposed framework is described in section II. Section III is devoted to the system performance evaluation. Finally, we conclude the paper and propose some directions for future research in section IV.

## II. SYSTEM DESCRIPTION

In this work, we aim to develop a hierarchical system to differentiate between driving maneuvers which are different in terms of longitudinal or lateral vehicle movement patterns. To this end, we propose two parallel learning based methods to predict vehicle trajectory in both directions, i.e. along with and perpendicular to the road lane direction. This overall system can be employed to either predict the host vehicle behavior, in which the system has been implemented, or any remote vehicle's trajectory, which are in the host vehicle's sensing or communication range. For instance, the latter case is illustrated in Figure 1. In either case, some specific applications would benefit the outputs of the proposed system. For example, host vehicle maneuver prediction is beneficial to applications such as lane keep assist system (LKAS) and blind spot warning (BSW). On the other hand, forecasting remote vehicle maneuver has a great effect on the performance accuracy of the other class of applications such as cooperative forward collision warning (FCW) and cooperative adaptive cruise control (CACC). The performance of the mentioned applications is directly affected by the accuracy of the maneuver prediction system. More precise prediction methods result in more accurate and smooth reactions of these applications to unforeseen and instantaneous driver's decisions. For instance, in CACC application which pursuits two main objectives, namely enforcing the vehicle to keep the safe distance from its leading vehicle and catching the closest possible velocity to what has been set by the driver, a better prediction of an abrupt cut-in maneuver gives the application more time to react and increases the overall system safety.

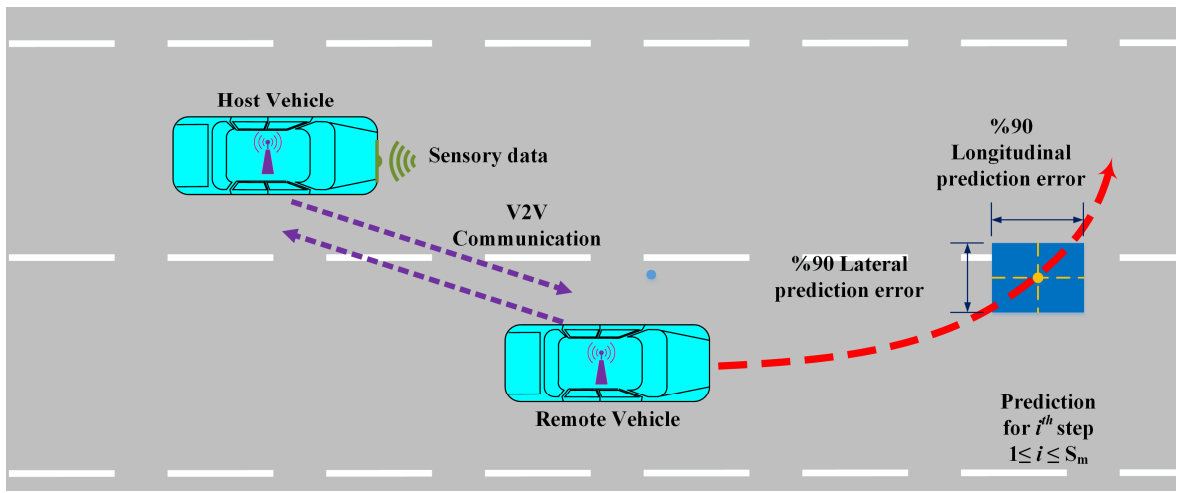

Figure 1- Remote vehicle trajectory prediction by the host vehicle

Our final objective to design the proposed system, as is stated above, is predicting the future maneuvers of the maneuvering vehicle during a reasonable time frame ahead. In other words, a high-level driving maneuver should be predicted by the model instead of only immediate vehicle kinematics. The length of this time frame should be selected appropriately in order to represent the direct consequences of the current driver decision. A very short prediction horizon does not have a great capability to enhance the application performance. On the other hand, a very long prediction duration could not be precise enough, as the effect of a probable change in driver decision appears beyond a certain time instant ahead, which invalidates our prediction for those moments based on his already observed behavior. We denote the optimum required prediction steps, which specifies a complete high-level maneuver and could be tuned based on the requirements imposed by the application, by $S_m$. According to some pioneering works in the literature [34], a 1-second window seems to be a reasonable time frame to capture the consequences of driver instantaneous decision. Therefore, in our settings $S_m$ has been set to 10, based on the default DSRC message broadcasting rate (10 Hz).

In general, there are two main sources of information which enable each vehicle in the system to achieve a level of situational awareness and then plan its future movements based on that. Cameras and on board detection devices such as radars and LiDARs are assumed as the primary information providers for automated and connected vehicle applications. In addition, Vehicle-to-Vehicle (V2V) communication, which is obtainable using DSRC devices, is regarded as an important supplementary information source whenever it is accessible. V2V communication facilitates a more precise decision making with insignificant additional cost. This technology provides the host vehicle with some specific selected Controller Area Network (CAN) bus parameters of the remote vehicles in a periodic broadcast manner. This set of parameters, which construct basic safety messages (BSM) [16], [33], is specified in BSM part one and part two of the SAE J2735 standard [33]. From BSM part one the following parameters are selected in this work as the input set, which is fed into our trajectory prediction system: latitude, longitude, elevation, speed, heading, steering wheel angle, 4-way acceleration set, and vehicle size. The latitude, longitude, and elevation represent the location of the vehicle's center of gravity in the WGS-84 coordinate system. The 4-way acceleration set consists of acceleration values in 3 orthogonal directions plus yaw rate, which are calculated based on the assumption that the front of the vehicle is toward the positive longitudinal axis, the right side of it defines the positive lateral axis, and a clockwise rotation results in a positive yaw rate.

A. *Proposed Trajectory Prediction Approach*

In general, both categories of information, i.e. local sensory data provided by CAN bus and received remote vehicle information via BSMs, in the case that V2V communication is enabled, are fed into our trajectory prediction system as its inputs. Positioning information should be converted to a rotated East-North-Up (ENU) coordinate system with respect to the position and heading of the host vehicle. The x and y coordinates represent the relative longitudinal and lateral positions of the maneuvering vehicle, respectively. A series of consecutively received values of each available parameter could be regarded as a time series. Different methods, such as autoregressive (AR), moving average (MA), autoregressive exogenous (ARX), hidden Markov model (HMM), nonlinear autoregressive (NAR), nonlinear autoregressive exogenous (NARX) and Artificial neural networks (ANNs), have been proposed in the literature to calculate the future values of time series based on its currently observed values. The best choice from these different methods depends on the different aspects of the modeled system. System behavior in terms of linearity is one of the most influential factors to select the appropriate prediction method. For instance, AR, MA or ARX methods are more applicable to linear time series. On the contrary, nonlinear systems could be modeled more precisely by other methods such as ANN. In general, different driving maneuvers impose nonlinear relations among raw available information values. This nonlinearity could be simply justified by the nonlinear nature of kinematics models as well-known and widely accepted driving maneuver modeling frameworks. Therefore, our prediction method is founded based on ANN as a widely accepted prediction tool for nonlinear systems description. More specifically, in the first step we use a nonlinear autoregressive (NAR), which utilizes a neural network to capture the nonlinear relation between its inputs and outputs, to predict the future $S_m$ steps of vehicle parameters. Then, a neural network-based nonlinear autoregressive exogenous (NARX) is proposed to predict the longitudinal trajectory. Finally, a recurrent neural network (RNN) is designed to predict the lateral vehicle trajectory. Feedback delay is a short term memory mechanism which is necessary for neural networks to capture the dependency of the next element of a time series to a finite set of its previous values and external inputs. All three aforementioned subsystems in this paper, i.e. NAR, NARX, and RNN, benefit from the feedback delay. Each of these three subsystems will be more elaborated later in this section. The overall proposed framework is depicted in Figure 2.

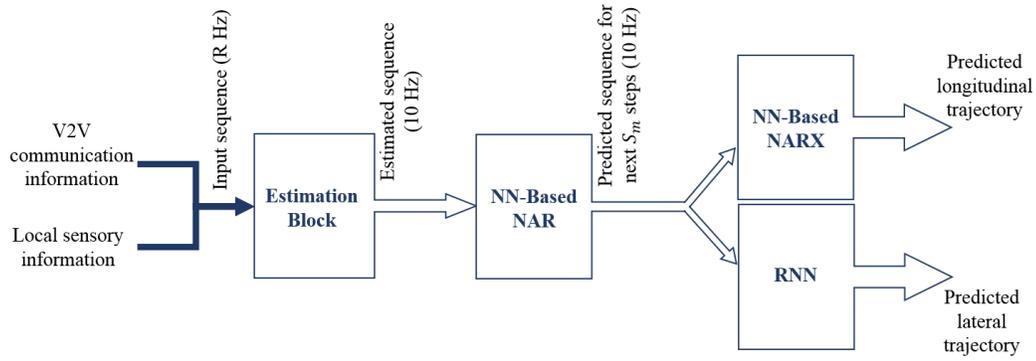

Figure 2- The proposed framework schematic

NAR is a modeling method which tries to find a nonlinear relation between the next value of a time series and its already observed values. This method does not rely on any exogenous inputs for its prediction. In this paper, a specific NAR is proposed to predict the future values of any of the input sequences to the vehicle system, namely steering wheel angle, yaw rate, heading, speed, and longitudinal acceleration. These sets of predicted parameters are needed to be fed into our next subsystems in order to calculate the future longitudinal and lateral trajectories.

In this paper, two different approaches are proposed to predict the longitudinal and lateral trajectories. Different subsets of the input sequences, which have been predicted using mentioned NAR block, are more critical for vehicle trajectory prediction along with each of these directions. This separation also enables our prediction system to be utilized, fully or partially, by different applications which are dependent on only longitudinal or lateral predicted trajectories or need both of them simultaneously. For instance, cooperative FCW needs only the longitudinal trajectory prediction, while lane keep assistance or blind spot applications rely on the trajectory prediction in both directions.

For longitudinal trajectory prediction, a NN-based NARX has been designed. In general, the output value of a NARX model depends on its past output values as well as a set of exogenous inputs. Some of the outputs of the previous step, i.e. predicted future values of the system input parameters, are assumed as the exogenous input for our longitudinal predictor. Therefore, the longitudinal motion of the vehicle is modeled using the predicted values of yaw rate, heading, speed, and longitudinal acceleration as external input signals to the system.

Finally, a RNN is adopted to derive the lateral trajectory prediction using the predicted values of steering wheel angle, yaw rate, and heading as its inputs. RNN is a class of ANNs in which the output of each hidden layer returns to that layer as a feedback signal. In spite of NAR and NARX which have finite input responses, RNN could have an infinite dynamic response. This infinite response is due to the mentioned recurrent feedbacks. Distinct high-level driver maneuvers with partially similar input time series could be differentiated by RNN by virtue of its infinite internal memory. For instance, a road curvature steering behavior is partially similar to the one from lane change maneuver. However, RNN could be trained to discriminate between these two maneuvers based on the other input signals.

Batch training is chosen for all these three ANNs due to the lack of sufficient data for online training and also its lower computational cost. Moreover, performing the training phase, which takes a long time and requires powerful computing, in an offline manner makes the proposed system more practical for our application, as the trained model would be ready to use by real time vehicular safety applications.

### III. EVALUATION

In order to evaluate the performance of the two proposed trajectory predictors along with longitudinal and lateral directions, a compound maneuver which includes movements in both mentioned directions would be preferable. These maneuvers allow the simultaneous assessment of the proposed framework performance in both directions. To this end, cut-in maneuver which is one of the most appropriate candidates, has been selected in this work. A number of cut-in maneuvers have been extracted from a set of realistic driving scenarios in Safety Pilot Model Deployment (SPMD) dataset [35], provided by National Highway Traffic Safety Administration (NHTSA) section of US Department of Transportation.

In general, any cut-in maneuver can be broken down into four consecutive phases, namely Intention phase, Preparation phase, Transition phase, and the Completion phase. Intention phase starts whenever a driver finds enough space between two successive vehicles in the adjacent lane and intends to change his lane. This phase does not incorporate any obvious physical actions. Only the overall circumstances are assessed to evaluate the possibility of a safe cut-in [36]. This driver assessment of the overall situation is subjective and extremely depends on the benefits of the cut-in, his driving style, his estimation of the vehicles formation in the adjacent lane and the required acceleration and speed for his maneuver [37]. In the next phase, i.e. preparation phase, driver adjusts the vehicle position in its current lane in terms of its longitudinal distance from the vehicles in the adjacent lane which are directly affected by cut-in maneuver. This adjustment is performed by proper longitudinal acceleration or deceleration. The third phase is conducted by a considerable lateral acceleration which is applied to shift the vehicle towards the adjacent lane. This phase continues with a lateral deceleration after the vehicle reaches its target lane. The lateral acceleration and lateral speed are bounded by the thresholds enforced by comfortable ride [38]. A smooth transition is achievable by bounding the lateral acceleration between -0.2g and 0.2g [39]. Finally, in the last phase, driver adjusts his speed in the new lane to hold a safe distance from its front and behind vehicles.

To demonstrate the superiority of our method, kinematics-based trajectory prediction is selected as a ground truth and the results of both methods on the same set of realistic cut-in maneuvers are presented. The kinematics model of the vehicle could be formulated as equations of (1) where $x_i$, $y_i$, and $v_i$ are longitudinal position, lateral position, and velocity of the $i^{th}$ vehicle, respectively. Also, $\phi_i$ denotes the steering angle, $\theta_i$ stands for the angle between the vehicle's instantaneous heading and the road direction, and $L_i = 5m$ is the length of the vehicle [40].

$$\begin{cases} \dot{x}_i = v_i \cos \theta_i \\ \dot{y}_i = v_i \sin \theta_i \\ \dot{\theta}_i = \frac{v_i}{L_i} \tan \phi_i \end{cases} \quad (1)$$

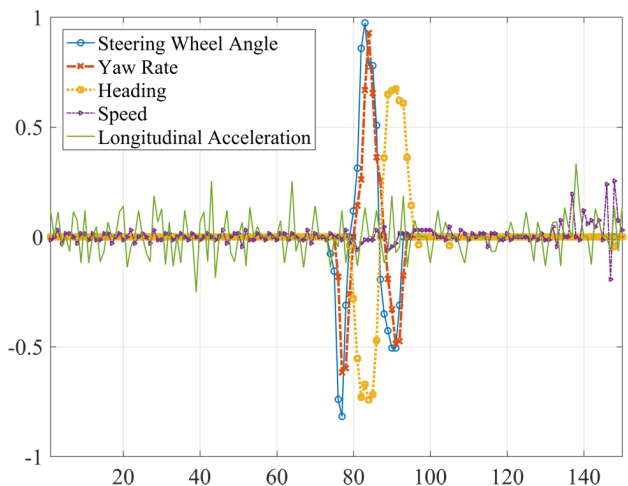

Figure 3- Smoothed, Normalized, and Integrated input signals of a single lane change maneuver

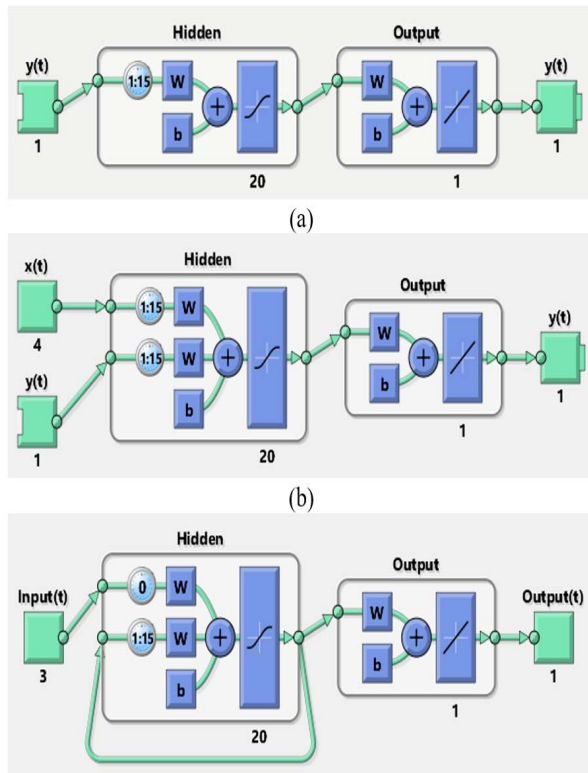

Figure 4- Layer structure of (a) NAR (b) NARX (c) RNN

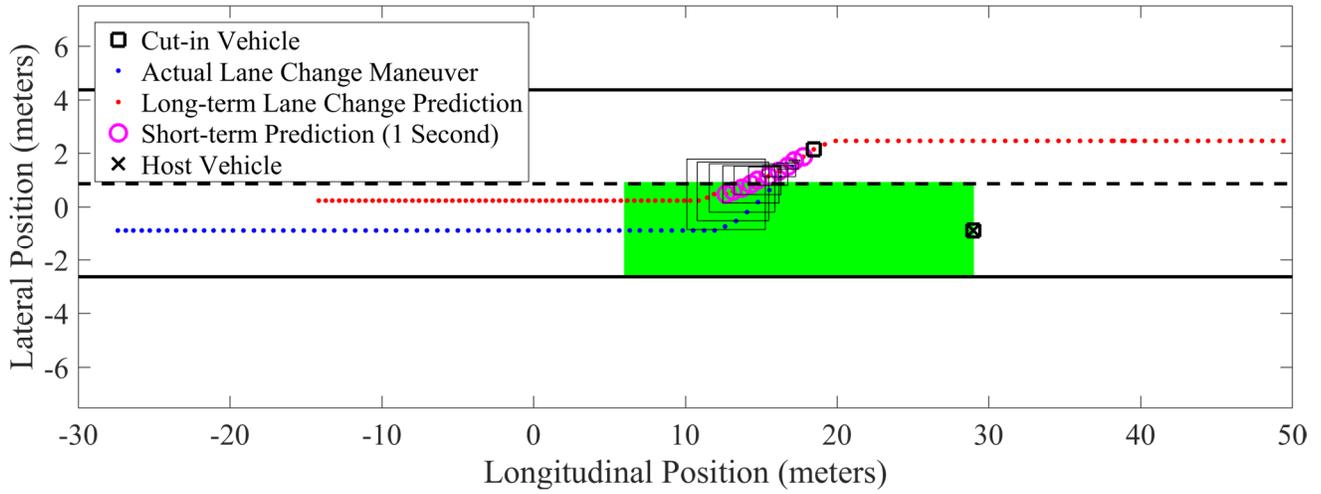

Figure 5- Joint perspective of longitudinal and lateral predictions

### A. Trajectory Prediction Performance Evaluation with Ideal Communication

For performance evaluation, 90 cut-in maneuvers have been extracted using the inherent information of received BSMs from participating vehicles in SPMD dataset in Ann Arbor, Michigan. In this section, we assume that the input sequences, which are received via both local sensors and communication, are available with the rate of 10 Hz. It means that an ideal channel for inter-vehicle communication is considered and packet drop effect is neglected.

The raw BSM information needs to be processed before using in our ANNs. In this preprocessing phase, the input signals for all ANNs are normalized to the range of $[-1,1]$ to increase the achieved performance. This is due to using Sigmoid as the activation function of each neuron. In addition, to reduce the linearity of the input signals or equivalently enhance the nonlinearity prediction process, differentiated sequence of input signals are used instead of the original ones. This new time series are known as integrated time series. The reconstruction of the predicted location values from what the ANNs return as a difference between any two consecutive values in the time series is performed by adding the first actual value to the sequence of the returned estimated differences.

Subsequently, to smooth the input time series and mitigate the effect of noise on them, we remove their small variations. More specifically, variations smaller than 3 degrees, 0.1 rad, 0.1 m/s, and 0.1 m/s$^2$ are removed from steering wheel angle, heading, speed, and longitudinal acceleration, respectively. These input signals for one of the analyzed maneuvers are illustrated in Figure 3.

As depicted in Figure 4, all three ANNs have 20 nodes in their hidden layers and 15 step short term memory, which is equivalent to use the past information of 1.5 seconds for future prediction due to assumed 10 Hz frequency. These values have been selected in our simulations based on the tradeoff between model complexity and its real-time applicability. Moreover, we set $S_m = 10$ to predict the vehicle trajectory for 1 seconds ahead. As mentioned before, predicting beyond this time frame is not accurate enough because driver might change his decision.

The dataset has been divided into three training, cross-validation, and testing sections with 70, 15, and 15 percent of the total available scenarios, respectively.

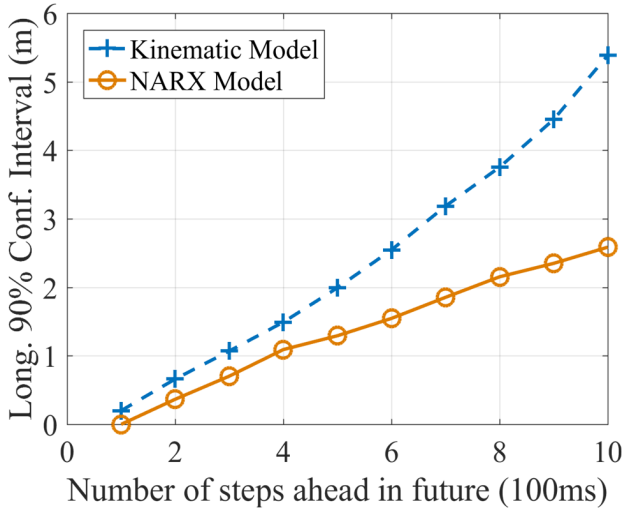

Figure 6- Comparison of 90-percentile conf. interval of longitudinal position prediction of the Kinematic and NARX models for different prediction steps

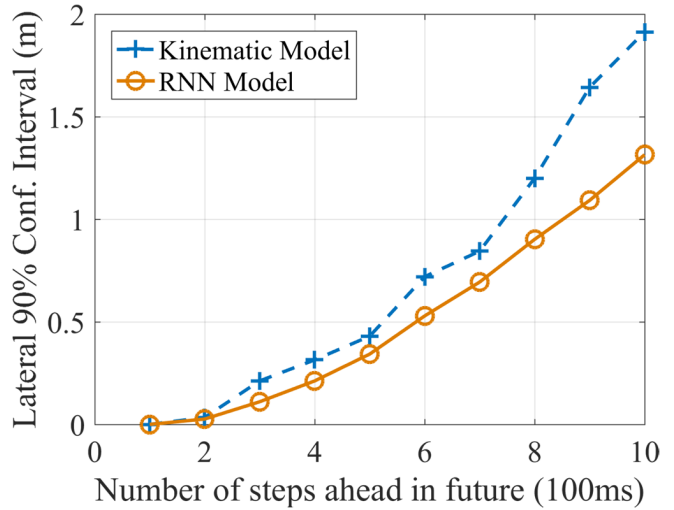

Figure 7- Comparison of 90-percentile conf. interval of lateral position prediction of the Kinematic and RNN models for different prediction steps

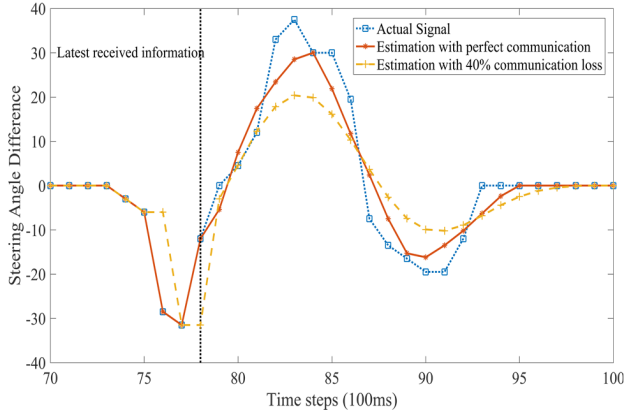

Figure 8 – Steering angle difference estimation comparison of ideal communication vs. 40% packet drop

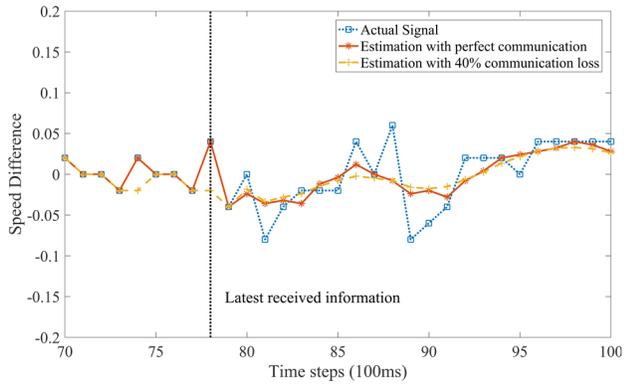

Figure 9- Speed difference estimation comparison of ideal communication vs. 40% packet drop

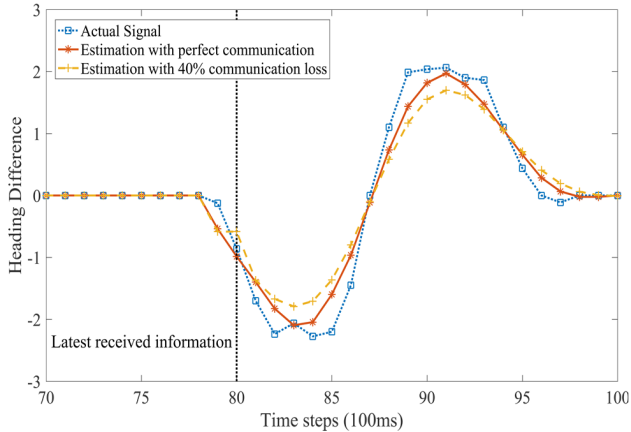

Figure 10 - Heading difference estimation comparison of ideal communication vs. 40% packet drop

The layer structure of our designed NAR, NARX, and RNN are depicted in Figure 4 (a), (b), and (c), respectively.

Combination of the longitudinal and lateral predictions, for one second ahead of a cut-in scenario from SPMD dataset, is depicted in Figure 5, for more clarification. In this figure, prediction errors are shown with consecutive rectangles on the predicted path. The performance of the prediction methods, i.e. vehicle kinematics model and our model (trained RNN and NARX), are compared using their 90 percentile accurate predictions for each of the 10 prediction steps. For both predictors, the averaged confidence levels for longitudinal trajectory prediction over all 90 scenarios are shown in Figure 6. Similarly, Figure 7 shows the same comparison for lateral position prediction versus kinematics model.

*B. Effect of Non-Ideal Communication on Parameter Predictions*

Communication network has been assumed as an ideal medium without any packet loss, so far. In this section, the effect of communication imperfection, in terms of packet loss rate, on the parameter sequence prediction is studied. To this end, the output of NAR prediction block for steering angle, speed and heading, for a communication network with 40% random packet drop rate is compared with their counterparts in an ideal network. In this work, a zero hold estimation method is utilized to reconstruct the received parameter sequences which are suffered from the packet loss. The results for steering angle, speed, and heading, for one of the analyzed cut-in maneuvers are depicted as an example on Figure 8-Figure 10, respectively.

The comparison of the estimated sequences shows a considerable similarity between our NAR block results in two different examined situations, i.e. ideal and 40% loss networks. This similarity, which is obtained using one of the simplest estimation methods, i.e. zero hold estimation, could be interpreted as the robustness of the proposed NAR prediction method against network imperfections. Employing more complicated estimation techniques would definitely enhance the prediction performance which could be considered as a future framework improvement.

## IV. CONCLUSION

In this paper, longitudinal and lateral vehicle trajectory predictions using local sensor and communicated information have been investigated. A novel multi-layer ANN-based framework is proposed to derive the future steps of either host or remote vehicle paths. In the first layer the incomplete sequence of received information due to network communication problems or sensor failures are reconstructed using zero-order estimation. Next, a NN-based NAR is utilized to generate the predicted sequence of the vehicle parameters for the duration of our trajectory prediction horizon ($S_m$). Finally, using the outputs of the previous step, two parallel neural network methods, i.e. NN-based NARX and RNN, are employed to predict the longitudinal and lateral trajectories, respectively. The whole proposed framework is evaluated on a set of realistic cut-in scenarios from SPMD dataset for both ideal and non-ideal communication network situations. The results demonstrate our method dominance over kinematics-based deterministic models as one of the most widely used methods in automotive industry.

Design of proper controllers to enhance the performance of different specific safety applications based on this trajectory prediction could be considered as an important future research topic.